\pdfoutput=1
\documentclass[11pt]{article}

\usepackage{ACL2023}

\usepackage{times}
\usepackage{latexsym}
\usepackage{amsmath}
\usepackage{amssymb}

\usepackage{booktabs}
\usepackage{colortbl}
\usepackage{multirow}
\usepackage{multicol}
\usepackage{pifont}
\usepackage{graphicx}
\usepackage{bm}
\usepackage{color, colortbl}
\usepackage[all]{nowidow}
\definecolor{Goldenrod}{rgb}{0.8, 0.6, 0.0}

\definecolor{Gray}{gray}{0.93}
\newcommand{\midsepremove}{\aboverulesep = 0mm \belowrulesep = 0mm}
\makeatletter
\newcommand*\bigcdot{\mathpalette\bigcdot@{.7}}
\newcommand*\bigcdot@[2]{\mathbin{\vcenter{\hbox{\scalebox{#2}{$\m@th#1\bullet$}}}}}
\makeatother

\newcommand{\xdownarrow}[1]{%
  {\left\downarrow\vbox to #1{}\right.\kern-\nulldelimiterspace}
}

\usepackage[T1]{fontenc}
\usepackage[utf8]{inputenc}

\usepackage{microtype}

\usepackage{inconsolata}

\newcommand{\graph}{\mathcal{G}}
\newcommand{\graphtr}{\mathcal{G}^{\textsc{tr}}}

\newcommand{\entities}{\mathcal{E}}
\newcommand{\entity}[1]{e_{#1}}
\newcommand{\entityset}{\mathbf{e}}

\newcommand{\relations}{\mathcal{R}}
\newcommand{\relationstype}[1]{\relations_\text{#1}}
\newcommand{\rel}[1]{r_{#1}}

\newcommand{\facts}{\mathcal{F}}
\newcommand{\factstype}[1]{\facts_{\text{#1}}}
\newcommand{\fact}[1]{f_{#1}}

\newcommand{\video}{v}
\newcommand{\videos}{\mathcal{E}_\video}
\newcommand{\pose}{v_p}

\newcommand{\phonset}{\boldsymbol{\phi}}
\newcommand{\phon}{\phi}
\newcommand{\phonemes}{\mathcal{E}_\Phi}

\newcommand{\semfeat}{\sigma}
\newcommand{\semfeatset}{\boldsymbol{\sigma}}
\newcommand{\semfeats}{\mathcal{E}_\sigma}

\newcommand{\sign}{s}
\newcommand{\signseq}{S}
\newcommand{\aslvocab}{\mathcal{E}_{\textsc{asl}}}
\newcommand{\enword}{w}
\newcommand{\envocab}{\mathcal{E}_{\textsc{en}}}

\newcommand{\topic}{t}

\defcitealias{aslphono}{de Amorim et al., 2022}
\defcitealias{liwc}{Tausczik et al., 2010}

\title{The American Sign Language Knowledge Graph: \\ Infusing ASL Models with Linguistic Knowledge }

\author{Lee Kezar \\
    Univ. of Southern California \\\And
  Nidhi Munikote \\
  Univ. of Southern California \\\And 
  Zian Zeng \\
  Univ. of Hawaii \\\AND
  Zed Sehyr \\
  Chapman University \\\And 
  Naomi Caselli \\
  Boston University \\\And 
  Jesse Thomason \\
  Univ. of Southern California}

\begin{document}
\maketitle
\begin{abstract}
% Draft 2
Language models for American Sign Language (ASL) could make language technologies substantially more accessible to those who sign.
To train models on tasks such as isolated sign recognition (ISR) and ASL-to-English translation, datasets provide annotated video examples of ASL signs.
To facilitate the generalizability and explainability of these models, we introduce the American Sign Language Knowledge Graph (ASLKG), compiled from twelve sources of expert linguistic knowledge.
We use the ASLKG to train neuro-symbolic models for~3 ASL understanding tasks, achieving accuracies of 91\% on ISR, 14\% for predicting the semantic features of unseen signs, and 36\% for classifying the topic of Youtube-ASL videos.

% Draft 1
% Language models for American Sign Language (ASL) could make language technologies substantially more accessible to those who sign.
% To align with the needs of its users, ASL models must transparently use video data and adapt to individual, potentially unseen signing styles.
% End-to-end models trained on ASL video data are somewhat accurate on isolated sign recognition and ASL-to-English translation, but are severely limited with respect to explainability, adaptability, and scalability.
% We argue that the \todo{(rephrase) broader task of ASL comprehension} requires additional breadth of knowledge, such as the phonological and semantic properties of signs, learned through neuro-symbolic techniques.
% To facilitate research into this approach, we built the ASL Knowledge Graph (ASLKG) from $12$ linguistical knowledge bases connected through shared relationships.
% We find empirical evidence that \todo{(rephrase) knowledge-infused learning} is an effective method for isolated sign recognition ($|V|\approx 2700$) and two novel tasks: recognizing the semantic features of unseen signs and classifying the topic of Youtube videos in ASL.
% Models trained on ASLKG data are $91\%$ accurate at sign recognition, $31\%$ accurate at semantic feature recognition, and $36\%$ accurate at topic classification; all of which improve over end-to-end neural models.
% \todo{4-5 sent}
\end{abstract}

\section{Introduction}
Computational models of signing aim to improve access to language technologies by automatically understanding and producing a sign language \citep{slp}.
Recognizing that sign language models could benefit signing communities broadly, including tens of millions of deaf and hard-of-hearing people, there have been repeated calls for new technological resources to assist in sign language modeling \cite{60-60, includingsl}.
These calls strongly emphasize the fair and equitable treatment of those who sign, including respect for signing communities' autonomy, diversity, and right to privacy \citep{deafsafeai}; improving the transparency of how models use signing data \citep{fate}; and mitigating audist biases in research \citep{bias, ableism}.

\begin{figure}[t!]
    \centering
    \includegraphics[width=\linewidth]{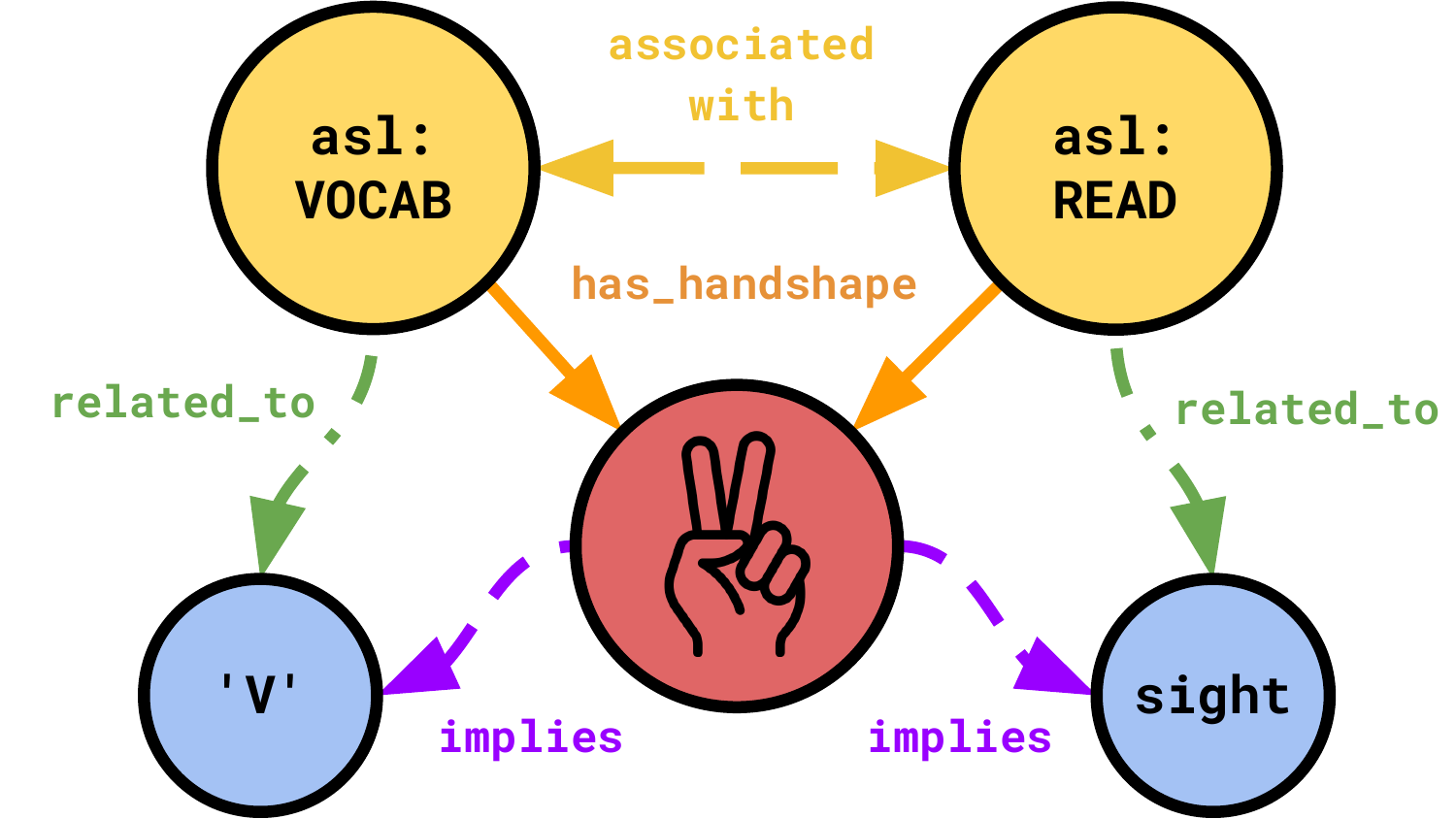}
    \caption{The ASLKG relates the \textcolor{red}{\textbf{form}} (e.g., \textit{2/V handshape}) and \textcolor{blue}{\textbf{meaning}} (e.g., related to \textit{sight}) of \textcolor{Goldenrod}{\textbf{signs}} in the ASL lexicon. We use this knowledge to neuro-symbolically recognize signs (e.g., \textsc{read}) and infer their meaning.} 
    % \textit{Note}: signs are more than a handshape, and the relationship between form and meaning is contextual and complex.}
    \label{fig:splash}
\end{figure}

In meeting these sociotechnical goals, sign language models must address the scarcity and inconsistent curation quality of signing data, including coverage of the lexicon, consistency of sign labels (if they exist at all), and documented representation among signers \citep{overview, glosscritique}.
% Taken together, methods for modeling sign languages must respect signing communities' values while learning the linguistic patterns of signing and appropriately handling variation (e.g., regional signs, accents, and dialects).
% In meeting the goals on tasks , the limitations of end-to-end modeling are clear.
Even using the largest and most externally valid datasets of American Sign Language (ASL) for isolated sign recognition (\textit{in}: video; \textit{out}: one of 2--3k sign identifiers), end-to-end, data-hungry neural models have struggled to surpass $70\%$ top-1 accuracy~\citep{asllvd, wlasl, citizen, semlex}.
% Additionally, it is not immediately clear how neural models arrive at their prediction, and the ability to adapt to an individual signer's style is not considered in the research.
Meanwhile, end-to-end models can translate ASL weather reports into English \citep{rwth} with a BLEU score of approximately $29$ \citep{translation-1, translation-2}; however, this metric does not clearly indicate where, how, and why models make mistakes.
% For example, it is unclear to what extent translation models correctly recognize signing \textit{in general}, especially the inflected and derived forms of root signs (e.g., \textsc{day} vs. \textsc{two-days}), neologisms, fingerspelling (e.g., \textsc{\#busy}), most numbers, classifier predicates (e.g., \textsc{cl:smell-candle}), indexing (e.g., \textsc{third-item}, as in a list), and more \citep{glosscritique}.

To facilitate technological progress on these fronts, we introduce the \textbf{American Sign Language Knowledge Graph} (ASLKG), a collection of over $71$k linguistic facts related to $5802$ ASL signs (§\ref{sec:kg}).
% Acknowledging that isolated sign data provide insufficient examples to reliably recognize, let alone understand these phenomena, we claim that a neuro-symbolic, knowledge-driven approach can improve ASL models' generalizability, data efficiency, explainability, and customizability.
% To support this claim, we introduce the \textbf{\mbox{American} Sign Language Knowledge Graph} (ASLKG), a collection of over $71$k linguistic facts related to $5802$ ASL signs.
The ASLKG is built from~$8$ knowledge bases pertaining to ASL linguistics, and supplemented with~$4$ more for English lexical semantics.
To validate the quality of the ASLKG, we use it to train neuro-symbolic models on three downstream tasks (§\ref{sec:method}): (a) recognizing isolated signs, (b) predicting the semantic features of out-of-vocabulary signs, and (c) classifying the topic of continuous ASL videos. 
% We call this proposed method \textit{linguistic knowledge infusion}, as we use expert-crafted facts related to the phonology, morphology, and semantics of ASL to interpretably reason about the structure and meaning of signs.
By grounding a video to phonological features in the ASLKG and reasoning about what those features might mean (e.g., signs, semantic features), we achieve $91\%$ accuracy at recognizing isolated signs, $14\%$ accuracy at predicting unseen signs' semantic features, and $36\%$ accuracy at classifying the topic of Youtube-ASL videos (§\ref{sec:results}).

The ASLKG is released under the CC BY-NC-SA 4.0 License at \href{https://drive.google.com/file/d/11zEtzoJG4a8XftEXKx9JLnOY0wAUrwI1/view?usp=sharing}{this link}.

\section{Background}

\subsection{Knowledge-Infused Learning} \label{sec:bg/neurosymb}
Neuro-symbolic methods combine data-driven pattern recognition with knowledge-driven reasoning over well-defined concepts \cite{thirdwave}.
These methods have been helpful in reasoning over linguistic patterns embedded in high-dimensional data, like video and audio \cite{nsnlp}.
By accurately grounding these patterns to abstract symbols, models can associate observations with various forms of expert knowledge \citep{neurosymb}.% and perform reasoning in a logical, interpretable way \citep{neurosymb}.

Knowledge infusion describes how expert knowledge informs the parameters of a neural model, and can improve performance and data efficiency~\citep{knowledgeinf}.
We apply the knowledge-infused learning framework \citep{kil} to improve models for sign recognition and understanding.
In these approaches, real-world observations are grounded to symbolic knowledge, such as a knowledge graph (KG), which represents facts in the form (\textit{subject}, \textit{predicate}, \textit{object}).\footnote{There is no widely agreed-upon definition of a knowledge graph. For further discussion, see \citet{kgdefine}.}
For instance, WordNet \citep{wordnet} encodes \textit{hypernym} relationships, such as (\textit{computer}, \textit{is-a}, \textit{technology}).

% After grounding video data to the graph, models may be trained to infer the existence of other relationships with quantifiable certainty.
In this paper, we ground isolated and continuous ASL videos to phonological features in the ASLKG (§\ref{sec:kg/tasks/grounding}), then \textit{infer} their corresponding signs and their semantic features (§\ref{sec:kg/tasks/inference}).
We additionally develop KG node embeddings, a form of knowledge-infused learning based on fact verification (§\ref{sec:kg/tasks/verification}), to include more holistic linguistic knowledge the inference process.

\subsection{ASL Linguistics} \label{sec:bg/ling}
Sign languages are complete and natural languages primarily used by deaf and hard-of-hearing people. 
Sign languages and spoken languages are similar in many ways: there are over one hundred distinct sign languages used by communities around the world \citep{eberhard_ethnologue_2023}; and sign languages demonstrate full phonological, lexical, and syntactic complexity \citep{padden_interaction_2016, liddell_american_1980, padden_native_2001, coulter_current_2014}.
Despite these broad similarities, sign languages also differ from spoken languages both in terms of articulation modality (\textit{visual-manual}) and phonology.
These differences necessitate specialized techniques for NLP.
The ASLKG focuses specifically on American Sign Language (ASL), a sign language used by communities in the United States and Canada.

\subsubsection{Phonology}
Understanding the phonological structure of sign languages has direct implications for computational models \citep{handposeguided, bsl1k, wlasl-lex, eacl}.
In this work, we adopt a phonological description of signs as a means of grounding video data to ASLKG (§\ref{sec:kg/tasks/grounding}).
Phonemes---the smallest units of language---constitute an inventory of visuo-spatial patterns that can be combined to create words, and are generally not considered to carry meaning.
In ASL, signs are distinguished phonologically based on the shape, orientation, movement, and location of the signer's hands, in addition to non-manual markers such as lip shape, eyebrow height, and body shifting \citep{herrmann_nonmanuals_2011, michael_framework_2011}.
These facets are sometimes referred to as \textit{sign language parameters}, and all signs take one or more values for each facet.

Unlike spoken languages, in sign languages, phonological parameters act as \textit{simultaneous} channels of visuo-spatial articulation.
For example, within one sign, the hands have a particular shape and are also in a particular location.
Signs also exhibit sequentiality, such as using multiple locations or handshapes.
Therefore, conceptualizing language as a sequence of linguistic items is not entirely accurate in the case of signing, adding modeling complexity.

\subsubsection{Lexical Semantics}
As an early step towards computationally modeling the relationships between phonological form and meaning in ASL, we associate signs with lexico-semantic features.
Traditionally, linguistic theory has asserted that meaning is conveyed through arbitrary symbols \citep{locke, saussure} and their co-occurrence \citep{firth}.
One of the unique affordances of the sign language modality, however, is that signs' phonological forms often physically resemble their meanings \citep{iconicity}, and these non-arbitrary associations between phonological form and meaning occur in patterned, systematic ways across the lexicon \citep{universals}. 

For example, \citet{borstell2016} found that across ten sign languages, signs representing multiple entities (e.g., \href{https://asl-lex.org/visualization/?sign=shoes}{\textsc{shoes}}, \href{https://asl-lex.org/visualization/?sign=family}{\textsc{family}}) often use two-handed signs.
Similarly, \citet{occhino2017} showed that in both ASL and Libras,\footnote{Also known as Brazilian Sign Language} convex and concave shapes (e.g., \href{https://asl-lex.org/visualization/?sign=ball}{\textsc{ball}}, \href{https://asl-lex.org/visualization/?sign=bowl}{\textsc{bowl}}) are frequently represented by signs using a claw handshape. These non-arbitrary forms do not always come about through productive or discrete processes \citep{systematicity}---not all plural signs are two-handed, and clawed handshapes do not exclusively denote rounded shapes.
Instead, these ``systematic'' forms are more fluid (i.e., subject to a degree of variation) while remaining preconditioned on meaning \citep{padden}. 
We posit that these flexible and pervasive mappings between form and meaning in sign languages are an important source of information that might improve models' ability to understand out-of-vocabulary signs.
We are able to use the ASLKG to confirm that the systematicity of ASL enables \textit{reasoning about the meaning of an unseen sign based on its form}.
% These features are treated as labels associated with either ASL signs or English words, e.g. \textsc{flower} is categorically related to the idea of \texttt{nature} (an example of \textit{hypernymy}).
% We assume that a subset of these semantic features will correlate with signs' phonological forms across the lexicon, and can therefore be modeled probabilistically.
% We further describe these features in Section \ref{sec:kg/definition}.

\begin{table*}[ht]
    \centering
    \midsepremove
    \begin{tabular}{llrrrccc}
        
        \toprule
         Dataset & Language & $|\text{Vocab}|$ & Phon. & Morph. & Syn. & Sem. \\
         \midrule
         ASL-LEX 2.0 \citep{asllex}                     & ASL, En.  & 2,723     & \checkmark & \checkmark & \checkmark & \\
         
         \rowcolor{Gray} ASLLVD \citep{asllvd}          & ASL, En.       & 3,314     & \checkmark & \checkmark & \checkmark & \\

         Associations \citep{semassoc}                  & ASL       & 3,149     & & & & \checkmark \\
         
         \rowcolor{Gray} IPSL \citep{iconicity_db}    & ASL       & 79        & \checkmark & \checkmark & & \checkmark \\
         
         Sem-Lex \citep{semlex}                         & ASL       & 3,149          & \checkmark & & & \\
         
         \rowcolor{Gray} ASL Citizen \citep{citizen}    & ASL       & 2,731           & & & & \\
         
         ASL Phono \citepalias{aslphono}                     & ASL       & 2,745          & \checkmark & & & \\
         
         \rowcolor{Gray} Youtube-ASL \citep{ytasl}      & ASL, En.  & ---  & & & \checkmark & \checkmark \\
         
         \midrule
         
         Lancaster \citep{lancaster}                    & En.       & 39,707 & &  & & \checkmark \\
         
         \rowcolor{Gray} WordNet \citep{wordnet} & En.  & 155,327   & & \checkmark & \checkmark & \checkmark \\
         
         LIWC \citepalias{liwc}                              & En.       & 12,000 & & \checkmark & \checkmark & \checkmark \\
         
         \rowcolor{Gray} Empath \citep{empath}          & En.       & 59,690 & & \checkmark & & \checkmark \\
         
         \midrule
         
         ASLKG                                          & ASL, En.  & 8,240 & \checkmark & \checkmark & \checkmark & \checkmark \\
         \bottomrule
    \end{tabular}
    
    \caption{The ASLKG brings together eight sources of knowledge pertaining to the linguistic structure ASL signs, with four English-based sources to supplement morphological, syntactic, and semantic facets via sign-level translation. (\textit{Key: Phon. = phonology, Morph. = morphology, Syn. = syntax, Sem. = semantics})}
    \label{tab:sources}
\end{table*}

\section{The ASL Knowledge Graph}
\label{sec:kg}
We define the elements and structure of the ASLKG (§\ref{sec:kg/definition}), report its statistics (§\ref{sec:kg/stats}), and describe how it can be used in machine learning tasks (§\ref{sec:kg/format}).
Then, we present three classes of ASLKG tasks (§\ref{sec:kg/tasks}): verifying whether new knowledge can be added to the graph (§\ref{sec:kg/tasks/verification}), grounding video data $\video$ to entities $\entities$ (§\ref{sec:kg/tasks/grounding}), and inferring implicit relationships from observations (§\ref{sec:kg/tasks/inference}).

\subsection{Definition} \label{sec:kg/definition}
The ASLKG $\graph$ is a set of entities $\entities$ interconnected by a set of relations $\relations$ according to a set of facts $\facts$ pertaining to ASL signs. Formally,
\begin{align}
    \graph &:= \langle \entities, \relations, \facts \rangle \\
    \entities &:= \{ \entity{1}, \entity{2}, \dots \entity{E} \}\\
    \relations &:= \{ \rel{1}, \rel{2}, \dots \rel{R} \} \\
    \facts &:= \{ \fact{1}, \fact{2}, \dots \fact{F} \}, \\
    \fact{} &\in \entities \times \relations \times \entities
\end{align}

Entities primarily cover ASL signs $\sign \in \aslvocab$, examples of those signs $\videos$ ($n=174547$), English words $\enword \in \envocab$, ASL phonemes $\phon \in \phonemes$, and ASL semantic features $\semfeat \in \semfeats$.
A number of numeric features, such as the number of morphemes or video duration, are also included in $\entities$.

\textbf{ASL signs} ($n=5802$) and \textbf{English words} ($n=2438$) are associated through expert-annotated labels, constituting a many-to-many relationship between $\aslvocab$ and $\envocab$.
For ASL signs, we additionally include video observations $\videos$ from Sem-Lex \citep{semlex} ($n=91148$) and ASL Citizen \citep{citizen} ($n=83399$), which have been manually labeled from a shared vocabulary $V \subset \aslvocab$ ($|V|=2723$).

The \textbf{phonological features} $\phonemes$ ($n=196$) describe patterns in articulation related to the hands, face, or body.
Through the ASL-LEX dataset~\citep{asllex}, $\videos$ are manually labeled with the phonological features and sign identifiers, enabling machine learning models for phonological feature recognition \citep{eacl, esann, 3dlex}.
In Section \ref{sec:kg/tasks/grounding}, we describe how these models may be useful for grounding video data to the graph.

The \textbf{semantic features} $\semfeats$ ($n=319$) describe patterns in meaning, such as semantic associations \citep{semassoc} and hypernym (is-a) relationships.
We supplement the ASL-based features with semantic features of English words ($n=312$) collected from sources like WordNet, LIWC, and Empath \citep{wordnet, liwc, empath}.
Collectively, the semantic features describe the meaning of lexical items \textit{a priori} at varying levels of abstraction.
Given the systematic relationships between phonology and semantics in ASL, these semantic features are potentially helpful in cases where a sign is out-of-vocabulary, such that the \textit{components} of that sign (e.g. a subset of the phonemes) systematically and partially signal the sign's meaning.

\subsection{ASLKG Statistics} \label{sec:kg/stats}
Altogether, the $\graph$ contains $22,931$ entities and $160$ relations distributed across $71,768$ facts.
Table \ref{tab:stats} shows that ASL is more represented than English in terms of unique entities and number of facts.
In general, facts usually take a lexical item as the subject and a descriptor as the object, therefore the average out-degree is greater than the average in-degree.
The high standard variation suggests that the knowledge is not evenly distributed across lexical items, aligning with the expectation that more frequent signs are more likely to appear in linguistic data generally.

\subsection{Partitioning Facts for ML Tasks}  \label{sec:kg/format}
% Description of datasets and preparation.
% For ISR and SFR, we train models from isolated sign examples $\video_\sign$ in Sem-Lex \citep{semlex} and ASL Citizen \citep{citizen}.  
% By querying ASLKG with $\video_\sign$'s identifier, we can easily retrieve its sign identifier $\sign \in \aslvocab$, phonological features $\phonset \in \phonemes$, and semantic features $\semfeatset \in \semfeats$.  
To assist with experimentation, we assign each video example to an \textit{instance fold} ($0\leq i <5$) and a \textit{sign fold} ($0\leq i <10$).
The instance folds $\videos^{(i)}$ are such that $p(\aslvocab)$ is approximately equal across folds, i.e. each sign is equally represented in each instance fold.
The sign folds $\aslvocab^{(i)}$ are equally-sized partitions of $\aslvocab$ to facilitate tasks involving \textit{unseen} signs, such as semantic feature recognition.
For ISR, we use cross-validation on the instance folds; for SFR, on the sign folds.

\subsection{Tasks Facilitated by the ASLKG} \label{sec:kg/tasks}
% - overarching description of the tasks
We propose three initial tasks with practical applications in ASL modeling that leverage the ASLKG: fact verification (§\ref{sec:kg/tasks/verification}),  sign video grounding (§\ref{sec:kg/tasks/grounding}), and knowledge inference (§\ref{sec:kg/tasks/inference}).

 \subsubsection{Intrinsic Fact Verification} \label{sec:kg/tasks/verification}
\textit{Verification} is the task of estimating the existence of some unseen fact $p(\fact{}'~|~\graph)$.
Verification is commonly used as a pretraining task for creating KG embeddings, where a graph neural network $M_E(f)$ learns to discriminate between true and false facts by means of a scoring function $s(\mathbf{h}, \mathbf{r}, \mathbf{t})$, where $(\mathbf{h}, \mathbf{r}, \mathbf{t})$ are learnable vectors for the head, relation, and tail elements, respectively.
Trained in this way, the goal of KG embeddings is to capture the stable, cohesive relationships among entities.

In this work, we apply KG embeddings to subgraphs of $\graph$ such that the embeddings capture specific types of relationships (e.g. form, meaning), and experimentally test the extent to which they help with downstream inference tasks.

\subsubsection{Grounding ASL Videos to Phonemes} \label{sec:kg/tasks/grounding} 
Given video $\video$, let \textit{grounding} model $M_\entityset$ approximate the probability that $\video$ is associated one or more KG symbols $\entityset \subset \entities$.
Isolated sign recognition \citep{slp} may be described as a grounding task, provided that there exists an injective mapping from the ISR model's output classes $\mathcal{Y}$ to signs $\aslvocab$: $M_{\sign}(x;\theta) \approx p(\aslvocab | x)$.
However, any subset of $\entities$ can be the target of grounding. % provided there are sufficient data to train a model.
In this work, we explore grounding to phonological features $\phonemes \subset \entities$ (§\ref{sec:method/grounding}) as the first step in ASL-input tasks, like isolated sign recognition.

\subsubsection{Inferring Signs and their Meanings} \label{sec:kg/tasks/inference}
Given a set of grounded symbols $\entityset \subset \entities$, possibly with their associated probabilities $p(\entityset|x)$, KG \textit{inference} attempts to estimate the presence of some target, for example a novel fact $p(\fact{}'|\entityset) \cdot p(\entityset|x)$.
The benefits of inference in a symbolic medium include: (a) reduced pressure to acquire many training examples; (b) the ability to explain how the model computed its prediction; and (c) deterministic estimations of uncertainty \cite{neurosymb}.
Each of these benefits is relevant to sign recognition, where neural models are generally less accurate at recognizing signs on the long-tail \citep{semlex}, and users may want to customize their model or calibrate trust in its output.

\paragraph{Isolated Sign Recognition (ISR).}
Given a video $\video_\sign$ of a signer demonstrating one sign $\sign \in \aslvocab$, the model $M_\sign$ aims to estimate $p(\sign | \video_\sign)$.
$M_\textsc{s}$ may be implemented as probabilistic inference over phonological observations $\phonset \subset \phonemes$ as:
$$p(\sign | \video) \approx p(\sign|\phonset) \cdot p(\phonset|\video_\sign)$$
On the note of generalizability to new signs as well as flexibility to user preferences,~$p(\sign|\phonset)$ can be approximated from relatively few observations~$(\video_\sign, \phonset)$. % without loss of interpretability.
We compare a number of knowledge-infused methods for this task in Section \ref{sec:method/isr}.

\paragraph{Semantic Feature Recognition (SFR).} 
For sign fold $i$, given a video $\video_\sign$ of a signer demonstrating an unseen sign $\sign \in \aslvocab^{(i)}$, the model $M_\semfeat$ aims to approximate the semantic features $p(\semfeatset | \video_\sign)$.
As with $M_\sign$, in a neuro-symbolic setting, $M_\semfeat$ may be implemented as $p(\semfeatset|\phonset) \cdot p(\phonset|\video_\sign)$.
We position this task as a first attempt at zero-shot isolated sign understanding, emulating the likely scenario where a sign recognition model encounters an out-of-vocabulary (OOV) sign.
In application, such a model could act as a ``semantic back-off'': when $\text{max}(p(\sign | \video_\sign))$ is sufficiently low, we might decide that the sign is OOV and, in its place, use semantic features.
We explore this inference task, along with its inverse task $p(\phonset | \semfeatset)$, in Section \ref{sec:method/sfr}.

\paragraph{Topic Classification.}
Given a video $\video_\topic$ containing natural, sentence-level signing about genre or topic $\topic$, topic classification seeks to approximate $p(\topic | \video_\topic)$.
Compared to isolated sign data, $\video_\topic$ is more realistic and also more information-dense.
Using neuro-symbolic methods, we may approach this task as a sequence of independent transformations:
$$\video_\topic \xrightarrow{\hspace{0.5em} M_\phon\hspace{0.5em} } \phonset \xrightarrow{\hspace{0.5em} M_\sign\hspace{0.5em} } \signseq \xrightarrow{\hspace{0.5em} M_\topic\hspace{0.5em}} \topic,$$
where $\phonset, \signseq$ is a sequence of phonemes and signs, respectively, according to a sliding window over $\video_\topic$.
Such a model could, for example, facilitate searching over repositories of uncaptioned ASL video data on YouTube.

\begin{table}[th]
    \centering
    \midsepremove
    \begin{tabular}{lc|cll}
        \toprule
        \multicolumn{1}{c}{Statistic} & & & \multicolumn{1}{c}{ASL} & \multicolumn{1}{c}{English} \\
        \midrule
        \rowcolor{Gray} \# Sources & & & 8 & 4 \\
        \# Entities (E) & & & 5802 & 2438 \\
        \rowcolor{Gray} \# Facts (F) & & & 43513 & 17877 \\
        Avg. In-Degree & & & 2.19\ (9.4) & 1.56\ (0.9) \\
        \rowcolor{Gray} Avg. Out-Degree & & & 33.03\ (34.9) & 4.23\ (2.6) \\
        $\text{\# Sources per } \entity{}$ & & & 1.56 (0.6) & 1.99 (1.1) \\
        \bottomrule
    \end{tabular}
    \caption{ASLKG statistics by language (std. dev.).}
    \label{tab:stats}
\end{table}

\section{Method}
\label{sec:method}
We first describe how we collected and formatted the ASLKG data (§\ref{sec:method/curation}).
Then, to evaluate the ASLKG's practicality in downstream applications, we apply linguistic knowledge infusion toward three ASL comprehension tasks: isolated sign recognition (§\ref{sec:method/isr}), semantic feature recognition (§\ref{sec:method/sfr})), and topic classification (§\ref{sec:method/topic}).
We approach these tasks using the ideas of knowledge-infused learning, in particular by applying linguistic priors to the model architecture, training algorithm, and inference process.

\subsection{Knowledge Curation} \label{sec:method/curation}
We began construction with a search for candidate knowledge sources pertaining to ASL structure.
Only datasets that express (a) linguistic knowledge of signs, (b) examples of signing (isolated, sentence) or (c) linguistic knowledge of English words that are associated with an ASL sign were included as knowledge sources.
English data are included to weakly supervise the semantic relationships among signs that have an English translation.
This search resulted in 12 tabular datasets jointly representing ASL signs, English words, and their linguistic features (detailed in §\ref{sec:kg/definition}).
% For each dataset, we confirm that rows correspond to observations and columns correspond to properties/labels.
% For each column, we manually label the range of values with respect to data type (e.g. \textit{string}), linguistic type (e.g. \textit{phonological}), and vocabulary (where applicable).
% At this stage we also manually remove any columns that are not relevant to the general task of ASL modeling.
% Then, we identify which of these columns identifies the lexical item (except in Youtube-ASL, which is sentence-level).
After manually verifying the rows correspond to observations, we transform each row into facts where \textit{subject} entities are lexical items, \textit{relations} are column names, and \textit{objects} is the corresponding value.
See Appendix \ref{appendix:ontology} for further details on the integration process.

\subsection{Intrinsic Fact Verification}
As a pretraining task to produce ASLKG embeddings, we train graph neural networks $M_E$ to estimate $p(\fact{})$, where $\fact{}$ is either true (sampled from $\facts$) or false (randomly constructed such that $\fact{}' \notin \facts$).
We use two implementations of $M_E$, Trans-E \citep{transe} and DistMult \citep{distmult}.
We train the embedding models (implemented by \texttt{kgtk}\footnote{\url{https://kgtk.readthedocs.io/en/latest/}}) for 100 epochs using the subgraph of $\graph$ where the tail entities are lexical items $\aslvocab \cup \envocab$, phonemes $\phonemes$ or semantic features $\semfeats$.

\subsection{Grounding Video Data to ASLKG} \label{sec:method/grounding}
To ground video data to $\graph$, such as for sign recognition, we use the Sign Language Graph Convolution Network ($\textsc{slgcn}_\phon$) to approximate $p(\phonset|\video_\sign)$ following \citet{slgcn, esann}.
This model is $85\%$ accurate at recognizing $n=240$ phonological features, on average.
On the ISR task, for test instance fold $i$, we train $\textsc{slgcn}_\phon$ on instance folds $\neq i$ (and all $10$ sign folds).
On the SFR task, for test sign fold $j$, we train $\textsc{slgcn}_\phon$ on sign folds $\neq j$.
When removing a sign fold, we completely remove all the facts pertaining to those signs in the fold before training.

\subsection{Isolated Sign Recognition (ISR)} \label{sec:method/isr}
To estimate $p(\sign | \video_\sign)$, we compare several models: $\textsc{slgcn}_\sign$, $\textsc{fgm}_\sign$, $\textsc{knn}_\sign$, and $\textsc{mlp}_\sign$.
These models are designed to capture varying degrees of linguistic knowledge.
$\textsc{slgcn}_\sign$ is a neural baseline trained to predict $\video_\sign \rightarrow \sign$ directly.
Meanwhile, $\textsc{fgm}_\sign$ and $\textsc{knn}_\sign$ are formed from simple heuristics, namely co-occurrence and distance statistics.
$\textsc{mlp}_\sign$ maps embedded representations of $\phonset$ to $\sign$.

\subsubsection{Factor Graph Model} 
The factor graph model $\textsc{fgm}_\sign$ approximates $p(\aslvocab|\phonset)$ according to a partition or \textit{factorization} of $\phonemes$, expressed as:  
$$\prod_{z_i} p(\sign|z_i) \text{ s.t. } z_i \subset \phonemes \hspace{0.2em} \land \hspace{0.2em} \bigcup_{z_i} = \phonemes \hspace{0.2em} \land \hspace{0.2em} \bigcap_{z_i} = \emptyset.$$  
Factors are selected based on Brentari's Prosodic Model \citep{brentari}, grouping phonological features according to articulators (hand configurations), place of articulation (hand location in 3D space), and prosodic features (movements).  
We employ belief propagation with message passing (implemented via \texttt{pgmpy}\footnote{\url{https://pgmpy.org}}) to infer marginal probabilities across the factors, ensuring efficient computation of $p(\sign|\phonset)$.

\subsubsection{$k$-Nearest Neighbors}  
The $k$-nearest neighbors $\textsc{knn}_\sign$ model approximates $p(\sign_a | \phonset_b)$ based on the distance between $\sign_a$ (which is replaced with its ground-truth phonemes $\phonset_a$) and observations $\phonset_b$.  
The distance metric is defined as:  
$$d(\phonset_{a}, \phonset_{b}) = 1 - \frac{1}{16} \sum_{i = 0}^{16} \delta[\phonset_{a}^i = \phonset_{b}^i] \cdot p(\phonset_{b}^i | x).$$
The final prediction is determined by a majority vote among the nearest $k$ items in $\aslvocab$, using the minimum distance metric to resolve any ties.  

\subsubsection{Multilayer Perceptron}  
The multilayer perceptron $\textsc{mlp}_\sign$ approximates $p(\aslvocab|\phonset)$ by learning features for each input.  
Although less interpretable than other models, MLPs effectively represent many-to-one mappings in training data and could outperform non-parametric and exact inference methods given $\phonset$.  
The architecture consists of a randomly initialized embedding layer ($d=32$) to learn a representation of each phoneme, followed by three hidden layers of sizes $(64/128/256)$ and then a linear projection to the output (size $2723$).  
$\textsc{mlp}_\sign$ is trained for 100 epochs using cross-entropy loss and the Adam optimizer.

\subsection{Semantic Feature Recognition (SFR)} \label{sec:method/sfr}
To learn $p(\semfeats|\phonset)$, we use either $\textsc{mlp}_\semfeat$ or linear regression models $\textsc{reg}$.
Both architectures use a randomly initialized embedding layer ($d=32$) to learn a coherent representation of each phoneme.
Similarly to $\textsc{mlp}_\sign$, $\textsc{mlp}_\semfeat$ has three hidden layers of sizes $[64, 128, 256]$ and then a linear projection to the output (size $319$).

\subsection{Topic Classification} \label{sec:method/topic}
For topic classification, we use Youtube-ASL videos ($n=11k$), which we divided into 80\% train, 10\% validation, and 10\% test.
We first generate topics for each video based on their English captions $c \in C$, then apply a pipeline to each video resulting in (a) a multichannel sequence of phonemes (§\ref{sec:method/grounding}), (b) a sequence of signs and their embedding (§\ref{sec:method/isr}), and (c) a single semantic embedding.

\paragraph{Topic Generation.} To generate the topics, we use Latent Dirichlet Allocation (LDA) with $n_\topic = 10$ on the lemmas in $C$, weighted by TF-IDF.
We use \texttt{spaCy}\footnote{\url{https://spacy.io}} to perform tokenization and lemmatization, and \texttt{sk-learn}\footnote{\url{https://scikit-learn.org}} to perform LDA.
We then associate each video with its topic (e.g. \texttt{news}, \texttt{vlog}) as the topic classifier's final target.

\paragraph{Grounding.} To retrofit the phonologizer model $M_\phon(\video, \theta)$ to sentence-level data, we use a sliding window approach.
We divide each video into a sequence of windows according to a width ${W \in \{ 60, 30, 15 \}}$ and step ${\Delta_f \in \{15, 30\}}$ frames.

\paragraph{Inference.} We apply an isolated sign recognition model $M_\sign \in \{ \textsc{fgm}_\sign, \textsc{knn}_\sign, \textsc{mlp}_\sign \}$ to the predicted phonemes to estimate $p(\sign|\phonset)$.
We select the most probable sign for each window and form a sequence or \textit{gloss} $\signseq$.
Duplicate signs that are adjacent in the sequence and windows where no $p(\sign)>0.1$ are removed.

\paragraph{Embedding.} Next, we embed the sequence of signs $\signseq$ using \textsc{BERT} (uncased), implemented by HuggingFace.\footnote{\url{https://huggingface.co}}
\begin{align}
    E(\signseq) &= \textsc{bert}([ E(\sign_i)~\forall \sign_i \in \signseq]) \\
    E(\sign) &= \sum_{\enword_i \in t(\sign)}E_\textsc{bert}(\enword_i)*p(\enword_i|s),
\end{align}
where $t(\sign)$ is the set of translations for $\sign$ queried from $\graph$ and $p(\enword_i | s)$ is provided by ASL-LEX 2.0, also in $\graph$.
$E(\signseq)$ is a $d=768$ vector that we hypothesize will represent the high-level meaning of the ASL sentences in the video.

\paragraph{Topic Classification.} Finally, to evaluate the quality of $E(\signseq)$ with respect to topic classification, we train an $\textsc{mlp}_t$ and $\textsc{knn}_t$ to map $E(\signseq) \rightarrow t$.
The MLP model has one hidden layer $d=100$ and is trained for 50 epochs using a cross-entropy loss.
We compare the model performance to random guess and majority class baseline models.

\begin{table}
    \centering
    \midsepremove
    \begin{tabular}{c l | c}
        \toprule
        
        $\pose \rightarrow \bigcdot$ & \multicolumn{1}{c|}{$M_\sign$} & \textsc{acc} \\
        \midrule

        $\sign$ & $\textsc{slgcn}_\sign(\pose, \theta)$ & 0.64\\
        
        \rowcolor{Gray} $(\sign, \phonset)$   & $\textsc{slgcn}_{\sign,\phon}(\pose, \theta)$ & 0.66 \\

        \midrule
        
        & $ \textsc{fgm}_\sign(\hat{\phonset}, \graphtr)$ & 0.48 \\

        \rowcolor{Gray} & $\textsc{knn}_\sign(\hat{\phonset}, \graphtr)$ & 0.81 \\

        \hspace{0.2em} $\phonset \rightarrow \sign$ \hspace{0.2em} & $\textsc{mlp}_\sign(\hat{\phonset}, E_\theta(\graphtr))$ \hspace{0.2em} & \hspace{0.2em} 0.85 \hspace{0.4em} \\

        \rowcolor{Gray} & $\textsc{mlp}_\sign(\hat{\phonset}, E_\textsc{d}(\graphtr))$  & 0.86 \\

        & $\textsc{mlp}_\sign(\hat{\phonset}, E_\textsc{t}(\graphtr))$ & \textbf{0.92} \\
        
        \bottomrule
    \end{tabular}
    \caption{Top-1 accuracy (\textsc{acc}) on isolated sign recognition given pose $\pose$. For embeddings: $E_\theta \sim \mathcal{N}(0,1)$; $E_\textsc{d}$ is DistMult; and $E_\textsc{t}$ is Trans-E.}
    \label{tab:phon2sign}
\end{table}
\begin{table}[t!]
    \centering
    \midsepremove
    \begin{tabular}{ c c c@{\hspace{0.1cm}} c c@{\hspace{0.05cm}}|c@{\hspace{0.05cm}} c c }
         \toprule
        \multicolumn{2}{c}{ $\pose \rightarrow \phonset$} & & \hspace{0.1cm} $\phonset \rightarrow \sign$ \hspace{0.1cm} & & & \multicolumn{2}{c}{\textsc{acc}($\signseq \rightarrow \topic$)} \\
         \cmidrule{1-2} \cmidrule{4-4} \cmidrule{7-8}
         \hspace{0.2cm} $W$ \hspace{0.1cm}  & \hspace{0.1cm}  $\Delta_f$ \hspace{0.1cm}  & & $M_\sign$ & & & \hspace{0.1cm} $\textsc{mlp}_t$ \hspace{0.1cm} & \hspace{0.1cm} $\textsc{knn}_t$ \hspace{0.2cm} \\
         \midrule
         60 & 15 & & $\textsc{fgm}_\sign$ & & & 0.15 & 0.21 \\
         \rowcolor{Gray} 30 & 15 & & & & & 0.24 & 0.29 \\
         15 & 15 & & & & & 0.26 & 0.14 \\
         % \midrule
         \rowcolor{Gray} 60 & 30 & & & & & 0.19 & 0.29 \\
         30 & 30 & & & & & 0.15 & 0.21 \\
         \rowcolor{Gray} 15 & 30 & & & & & 0.21 & 0.37 \\
         \midrule
         60 & 15 & & $\textsc{knn}_\sign$ & & & 0.34 & 0.27 \\
         \rowcolor{Gray} 30 & 15 & & & & & 0.25 & 0.15 \\
         15 & 15 & & & & & 0.28 & 0.28 \\
         % \midrule
         \rowcolor{Gray} 60 & 30 & & & & & 0.34 & 0.25 \\
         30 & 30 & & & & & 0.25 & 0.30 \\
         \rowcolor{Gray} 15 & 30 & & & & & 0.24 & 0.23 \\
         \midrule
         60 & 15 & & $\textsc{mlp}_\sign$ & & & 0.24 & 0.25 \\
         \rowcolor{Gray} 30 & 15 & & & & & 0.34 & 0.28 \\
         15 & 15 & & & & & 0.24 & 0.15 \\
         % \midrule
         \rowcolor{Gray} 60 & 30 & & & & & \textbf{0.36} & 0.25 \\
         30 & 30 & & & & & 0.31 & \textbf{0.31} \\
         \rowcolor{Gray} 15 & 30 & & & & & 0.28 & 0.23 \\
         \bottomrule
    \end{tabular}
    \caption{Topic classification top-1 accuracy (\textsc{acc}) for $10$ topics. During $\pose \rightarrow \phonset$, window width $w$ and step $s$ are varied. Random guess on topic prediction is $0.14$ and majority class is $0.21$.}
    \label{tab:sign2topic}
\end{table}

% \begin{table*}[]
%     \centering
%     \begin{tabular}{c|cc|cc|cc}
%     \toprule
%          & \multicolumn{2}{c}{$w=15$} & \multicolumn{2}{c}{$w=30$} & \multicolumn{2}{c}{$w=60$} \\
%          & $s=15$ & $s=30$ & $s=15$ & $s=30$ & $s=15$ & $s=30$ \\
%     \midrule

%     \bottomrule
%     \end{tabular}
%     \caption{Caption}
%     \label{tab:my_label}
% \end{table*}

\section{Results}
\label{sec:results}
We report the results of our experiments on isolated sign recognition, semantic feature recognition, and topic classification.
In general, we find that the selected neuro-symbolic methods improve over comparable end-to-end techniques. 

\subsection{Isolated Sign Recognition}
On ISR, we report the top-1 accuracy across models and task configurations in Table \ref{tab:phon2sign}.
These results suggest that shallow knowledge infusion, operationalized as linguistic priors on $p(\sign|\video)$, improves over end-to-end models by $18.9\%$.
We additionally show that intrinsic fact verification, resulting in pretrained embeddings for $\phonset$, improves over end-to-end models by $25.2\%$.
The best model is $92\%$ accurate and therefore effective at ISR, a precursor to many ASL-input tasks.

\subsection{Semantic Feature Recognition}
On the novel task of SFR, we report $\textsc{f}_1$, precision, and recall in Table \ref{tab:sem_preds}.
As with ISR, intrinsic fact verification as a pretraining task improves over end-to-end models on $\phonset \rightarrow \semfeatset$ by $7$ points of accuracy.
We additionally find several semantic features that are recognized with relatively high F1: signs related to music ($\textsc{f}_1=0.63$), the body ($\textsc{f}_1=0.61$), and family ($\textsc{f}_1=0.50$).
The best model is $14\%$ accurate at recognizing semantic features, and in some cases may be useful at recovering from out-of-vocabulary signs.
For $\semfeatset \rightarrow \phonset$, knowledge infusion improves recognition accuracy by $11\%$ over end-to-end.
With further development, this latter task could be helpful in ASL-output tasks, where the intended meaning is already known, and a separate model can translate the phonological features into a coherent sign.

\subsection{Topic Classification}
We find that the LDA topics thematically align with the those reported in Youtube-ASL, including \texttt{vlogs}, \texttt{news}, \texttt{religion}, and \texttt{lessons} (\citet{ytasl}; ground truth topics were not released to the public).
In Table \ref{tab:sign2topic}, we report the top-1 accuracy with respect to window width $W$, step $\Delta_f$, ISR model $M_\sign$, and topic recognition model $M_t$.
We find that deep knowledge infusion, operationalized as a combination of grounding, inference, and KG embeddings, improves over a majority-class classifier by up to $15\%$.
The best model is $36\%$ accurate at classifying topics, and could assist in searching over large ASL corpora.

\begin{table}
    \centering
    \midsepremove
    \begin{tabular}{l c|c cc c cc}
        \toprule
        & & & \multicolumn{2}{c}{$\phonset \rightarrow \semfeatset$} & & \multicolumn{2}{c}{$\semfeatset \rightarrow \phonset$} \\
        \cmidrule{4-5} \cmidrule{7-8}
        \multicolumn{1}{c}{\multirow{-2}{*}{$M$}} & & & $\textsc{f}_1$ & \textsc{acc} & & $\textsc{f}_1$ & \textsc{acc}  \\
        \midrule
        $\textsc{mlp}[E_\theta( \bigcdot)]$ & & & 0.05 & 0.07  & & 0.18 & 0.20  \\
        $\textsc{reg}[E_\theta( \bigcdot)]$ & & & 0.01 & 0.02 & &  0.14 & 0.15  \\

         \rowcolor{Gray} $\textsc{mlp}[E_T( \bigcdot)]$ & & & 0.08 & 0.08 & & \textbf{0.28} & \textbf{0.31}  \\
        \rowcolor{Gray} $\textsc{reg}[E_T(\bigcdot)]$ & & & 0.04 & 0.05 & & 0.23 & 0.25  \\

         $\textsc{mlp}[E_D(\bigcdot)]$ & & & \textbf{0.12} & \textbf{0.14} & & 0.25 & 0.27  \\
         $\textsc{reg}[E_D( \bigcdot)]$ & & & 0.07 & 0.09 & &  0.21 & 0.22 \\
         \bottomrule
    \end{tabular}
    \caption{$\textsc{f}_1$ and accuracy (\textsc{acc}) on semantic feature recognition and the inverse task, semantic-to-phoneme recognition. (\textit{Key:} $E_\theta( \bigcdot) =$ \textit{randomly-initialized embeddings}; \textsc{reg} = \textit{linear regression model.})}
    \label{tab:sem_preds}
\end{table}

\section{Discussion}
In this work, we introduced the American Sign Language Knowledge Graph containing $71$k linguistic facts related to $5.8$k signs.
We show empirical evidence that the ASLKG is an effective resource for modeling American Sign Language input tasks.

On isolated sign recognition, we show that grounding video data to the graph and inferring the sign probabilistically is an accurate, scalable, and interpretable option for large sign vocabularies.
Additionally, we show that pretraining on fact verification to produce node embeddings adds an additional 1-7\% points of accuracy.

On semantic feature recognition, we show that unseen signs can be partially understood by mapping observed phonological features to semantic labels, such as ``related to family'', based on form alone.
On this task, a simple MLP model is 14\% accurate on average, also aided by node embeddings of the input.
Future work may explore more sophisticated methods for recognition or apply the recognized features towards understanding tasks.

On topic classification, we sequence grounding and inference models on sentence-level Youtube data, achieving an accuracy of $36\%$ at classifying from ten topics, achieving a $15\%$ improvement over majority class classifier.

As models for ASL attempt to overcome issues with data scarcity and curation quality, our results suggest that including expert-annotated linguistic knowledge through neuro-symbolic mechanisms is an effective path forward.
Future versions of the ASLKG will continue to refine the quality of the facts, add additional sources, and ship with more tools for knowledge-infused modeling techniques.

% \newpage

\subsection{Limitations} \label{sec:disc/limits}
The ASL lexicon is not fixed with respect to the signs in the lexicon, the way those signs are produced, or what those signs mean.
Variation exists at all levels of analysis, especially with respect to accent, dialect, and context.
These factors are not well-represented in ASLKG, because the primary focus of this work is establishing normative descriptions of ASL structure.

Excluding certain forms of signing disproportionately harms linguistic communities within ASL, such as those who use an underrepresented dialect.
We strongly discourage the use of ASLKG towards user-facing applications without the meaningful collaboration of ASL signers, especially those who are deaf and hard-of-hearing.\footnote{To locate potential collaboration with deaf and hard-of-hearing scholars interested in sign language technologies, consider the CREST Network: \url{https://www.crest-network.com}.}

\paragraph{Phonology}
Given the tremendous variation in sign language production across signers, despite including many signs in our grounding procedure, our approach  does not capture the phonological variation of ASL.
Additionally, our approach to discretizing ASL phonology represents only one way to divide an inherently non-discrete system, which is subject from ongoing debate from sign phonologists. 
For example, ASL-LEX 2.0 \citep{asllex} describes eight path movements, while SignWriting describes over 220 \citep{signwriting}. 
As we continue to refine the ASLKG, we may determine that certain parameterizations of sign phonology are best-suited for different target applications.

\paragraph{Lexical Semantics}
The use of English data to complement our linguistic knowledge of ASL assumes that there is sufficient overlap in the semantic structure of the two languages.
But, these are two independent languages with considerably different structures.
Although \textsc{BERT} is based on English grammar and semantics, we here assume that syntax does not play a significant role in capturing the topic of videos, and the meaning of an ASL sign is roughtly the meaning of its corresponding English gloss.
Both of these assumptions are standard approach in low-resource language modeling, but limit the fidelity of the representation.

\bibliography{anthology,custom}
\bibliographystyle{acl_natbib}

\appendix

\begin{figure*}
    \centering
    \includegraphics[width=\linewidth]{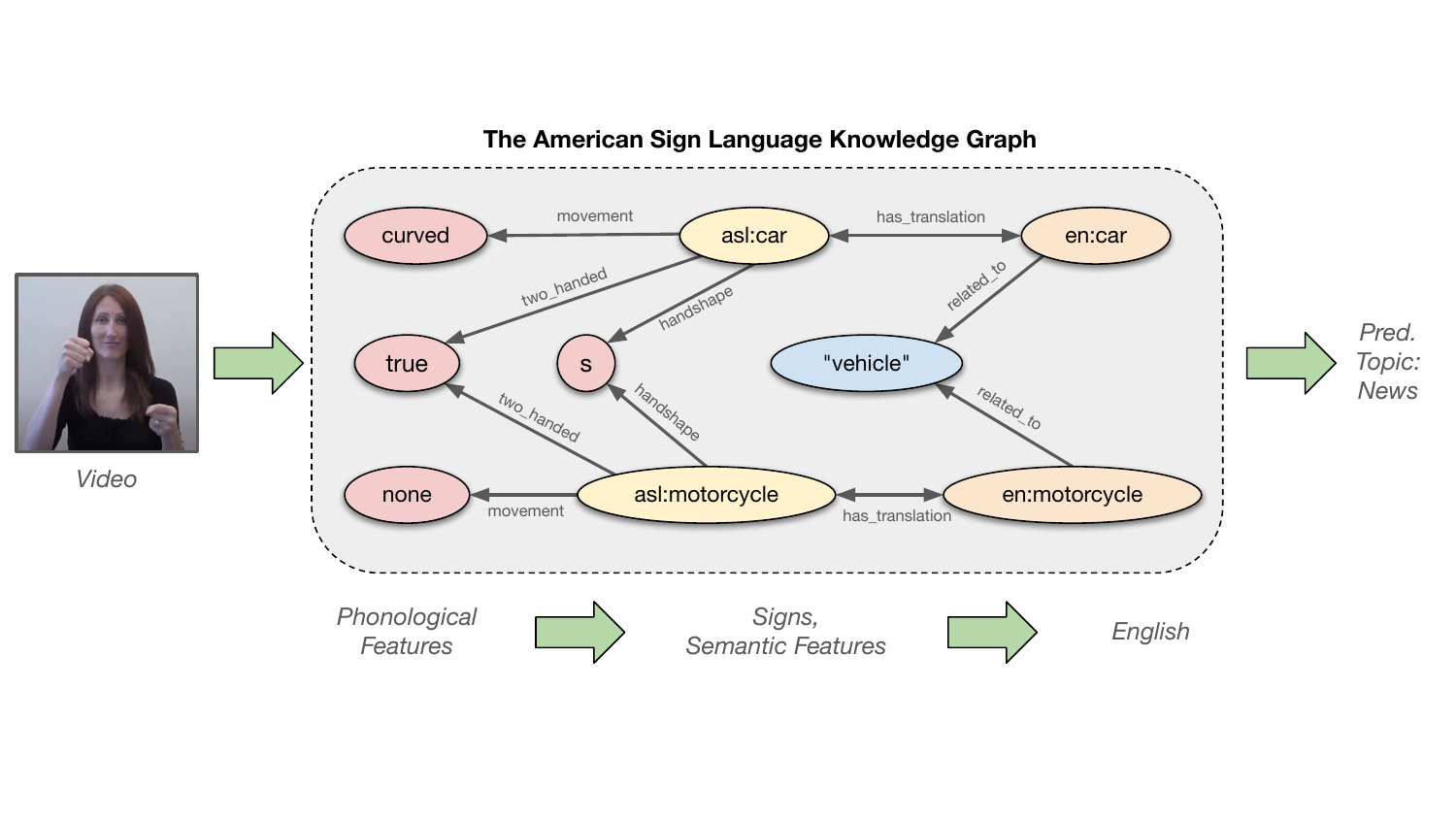}
    \label{fig:enter-label}
\end{figure*}

\section{Additional Statistics}
See Table \ref{tab:relation_types} for additional details regarding the types of knowledge included in the ASLKG.
% \begin{table*}[t!]
%     \centering
%     \begin{tabular}{l|lrrl}
%         \toprule
%         \multicolumn{1}{c|}{Relation Type $t$} & \multicolumn{1}{c}{Describes the subject sign's...} & \multicolumn{1}{c}{$|\relationstype{t}|$} & \multicolumn{1}{c}{$|\factstype{t}|$} & Example $\rel{} \in \relationstype{t}$ \\
%         \midrule
%         phonetic & sub-phonological production & 2 & 10,884 & \texttt{Sign\_Duration} \\
%         phonological & phonemes & 90 & 106,247 & \texttt{Handshape} \\
%         morphological & morphemes & 1 & \todo{5,729} & \texttt{Number\_Of\_Morphemes} \\
%         syntactic & lexical class/part of speech & 3 & 12,942 & \texttt{Lexical\_Class} \\
%         semantic & meaning (in isolation) & 285 & 23,963 & \texttt{Associated\_With}\\
%         translation & English translation & 14 & 28,925 & \texttt{Entry\_ID} \\
%         systematicity & form/meaning interaction & 27 & 27,895 & \texttt{Initialized\_Sign} \\
%         statistical & frequency & 30 & 98,007 & \texttt{Frequency\_N} \\
%         cognitive & mental representations & 2 & 2,132 & \texttt{Age\_Of\_Acquisition} \\
%         meta & meta-information & 2 & 6,714 & \texttt{SignBank\_Reference\_ID} \\
%         \bottomrule
%     \end{tabular}
%     \caption{The types of knowledge in ASLKG. $|\relationstype{t}|$ and $|\factstype{t}|$ denote the number of unique relations and facts, respectively, with type $t$.}
%     \label{tab:relation_types}
% \end{table*}

\begin{table*}[ht!]
    \centering
    \midsepremove
    \begin{tabular}{l|lrrl}
        \toprule
        \multicolumn{1}{c|}{Relation Type $t$} & \multicolumn{1}{c}{Describes the subject sign's...} & \multicolumn{1}{c}{$|\relationstype{t}|$} & \multicolumn{1}{c}{\hspace{0.8cm} $|\factstype{t}|$ } & Example $\rel{} \in \relationstype{t}$ \\
        \midrule
        phonetic & sub-phonological production & 2 & 2 733 & \texttt{Sign\_Duration} \\
        \rowcolor{Gray} phonological & phonemes & 90 & 94 218 & \texttt{Handshape} \\
        morphological & morphemes & 1 & 5 553 & \texttt{Number\_Of\_Morphemes} \\
        \rowcolor{Gray} syntactic & lexical class/part of speech & 3 & 5 657 & \texttt{Lexical\_Class} \\
        semantic & meaning (in isolation) & 285 & 26 581 & \texttt{Associated\_With} \\
        \rowcolor{Gray} translation & English translation & 14 & 28 925 & \texttt{Entry\_ID} \\
        systematicity & form/meaning interaction & 27 & 29 552 & \texttt{Initialized\_Sign} \\
        \rowcolor{Gray} statistical & frequency & 30 & 43 763 & \texttt{Frequency\_N} \\
        cognitive & mental representations & 2 & 16 282 & \texttt{Age\_Of\_Acquisition} \\
        \rowcolor{Gray} meta & meta-information & 2 & 9 429 & \texttt{SignBank\_Reference\_ID} \\
        \bottomrule
    \end{tabular}
    \caption{The types of knowledge in ASLKG. $|\relationstype{t}|$ and $|\factstype{t}|$ denote the number of unique relations and facts, respectively, with type $t$.}
    \label{tab:relation_types}
\end{table*}

% Updated values
% phonological     94218
% other            43763
% systematicity    29552
% semantic         26581
% cognitive        16282
% metadata          9429
% syntactic         5657
% morphological     5553
% phonetic          2733

\section{Refining the Ontology} \label{appendix:ontology}
The KG's ontology defines the space of possible entities $\entities$ and relations $\relations$.

To reduce the number of duplicate relations, we manually inspect each relation's meaning and values to determine if two columns are representing the same relationship.
We found two relations where this is the case in $\relations$: handshape and translation.
The handshape features included in ASLLVD largely overlap with those of ASL-LEX, but use slightly different name conventions.
We merged these two sets of handshape entities by manually renaming the ASL-LEX values to match the ASLLVD values.
For translation, we rename all relations to \texttt{has\_translation}, as it is already accepted that there is a many-to-many relationship between ASL signs and English words.

Likewise, there are pairs of elements in $\entities$ that refer to the same lexical item, and ought to be given the same label.
We merge the English vocabularies by replacing their prefix with \texttt{en:}, since the subject label uniquely and consistently identifies one lexical item.
English words that have no Levenshtein distance <= 1 to at least one of the signs' English translations are removed.

For ASL, however, there is no standard written component, so it is convention to label a sign with an English word called a \textit{gloss}, optionally including an ID for \textit{variations} (e.g. \texttt{asllex:right\_1} meaning ``correct'' and \texttt{asllex:right\_2} meaning the direction) or \textit{phonological features} (e.g. \texttt{asllrp:(1h)happy}) to help distinguish signs with the same gloss.
Therefore, additional support or evidence is necessary to justify combining two signs with the same label.
We use the handshape relation as a heuristic alongside Levenshtein distance to determine node equality:
$$\entity{a} = \entity{b} \text{ iff } \texttt{hs}(\entity{a}) = \texttt{hs}(\entity{b}) \land$$
$$\texttt{edit\_dist}\left[\texttt{gloss}(\entity{a}),\texttt{gloss}(\entity{b})\right] <= 1.$$

For all other non-word values, we remove NaNs, trailing zeros, and duplicate facts.

\end{document}